\documentclass{article} 
\usepackage[final]{acl}

\usepackage{hyperref}
\usepackage{url}

\usepackage{multirow}
\usepackage{booktabs}
\usepackage{natbib}
\usepackage{multirow}
\usepackage{tcolorbox}
\usepackage{tikz}
\tcbuselibrary{skins}
\usepackage{pifont}
\usepackage{caption}
\usepackage{subfigure}
\usepackage[graphicx]{realboxes}
\usepackage{wrapfig}
\tcbuselibrary{raster}
\usepackage{tabularx}
\usepackage{amsmath}
\tcbuselibrary{breakable}

\usepackage{wrapfig}
\usepackage{xspace}
\usepackage{enumitem}
\usepackage{xcolor}

\definecolor{darkpastelgreen}{rgb}{0.01, 0.75, 0.24}

\newcommand{\mname}{{\sc Pacit}\xspace}
\newcommand{\mnamec}{{\sc Pacit}}
\newcommand{\mnamet}{PACIT\xspace}

\title{\mnamet: Unlocking the Power of Examples for \\ Better In-Context Instruction Tuning}

\author{
  Tianci Xue$^1$\thanks{Work done during the internship at SUSTech.}, Ziqi Wang$^2$, Yixia Li$^3$, Yun Chen$^4$, Guanhua Chen$^3$\thanks{Corresponding author.} \\
  $^1$ Nanjing University, $^2$ University of Illinois Urbana-Champaigns \\
  $^3$ Southern University of Science and Technology\\
  $^4$ Shanghai University of Finance and Economics \\
  \texttt{xuetianci@smail.nju.edu.cn}, \texttt{ziqiw9@illinois.edu},   \texttt{liyixia@me.com} \\
  \texttt{yunchen@sufe.edu.cn},  \texttt{chengh3@sustech.edu.cn} \\
}

\begin{document}

\maketitle

\begin{abstract}
Instruction tuning enhances the instruction following ability of large language models by finetuning with supervised instruction data. Previous work proposes in-context instruction tuning (ICIT) where specific positive or negative examples are incorporated into the prompt for better performance. In this work, we propose PACIT, a simple and effective in-context instruction tuning method, inspired by the pedagogical concept of \emph{desirable difficulty}. The PACIT method unlocks the power of examples by encouraging the model to actively learn to grasp the distinctions between the positive and negative examples instead of merely reading. The model is expected to first verify the correctness of the provided example according to the task description, which is then set as the condition for generating a better response to the task instance. 
Our extensive experiments prove the effectiveness of \mname, outperforming ICIT baseline on both in-domain and out-domain tasks up to 9.16 and 3.14 average ROUGE-L scores, respectively. 
Moreover, PACIT can notably enhance the performance of instruction tuning even when all positive and negative examples are generated with a self-instruct method. 

\end{abstract}

\section{Introduction}
Large language models (LLMs) have garnered significant interest from both academia and industry due to their superior performance on a variety of natural language processing tasks such as question answering and text generation. Instruction tuning (IT; \citealt{ouyang2022training}) optimizes the pretrained language models with supervised instruction data to enhance the capabilities of the instruction following and zero-shot generalization to unseen tasks \citep{chung2022scaling,ouyang2022training,sanh2022multitask,alpaca,xue2023parameterefficient}. 
InstructGPT \citep{ouyang2022training} proposes in-context instruction tuning (ICIL) where the LLM is finetuned using instruction data with few-shot human-crafted positive examples. 
SuperNI \citep{wang2022supernaturalinstructions} presents a variant of in-context instruction tuning by further incorporating specified positive and negative examples in each task. The ICIL method achieves significant improvement compared with the vanilla zero-shot instruction tuning method \citep{ouyang2022training,wang2022supernaturalinstructions,li2023mimicit} with the knowledge from the demonstrations. 

However, previous in-context instruction tuning merely shows the specified positive and negative examples in the prompt, without further considerations for better digestion of examples. LLMs still struggle to follow the instructions precisely in some scenarios \citep{li2023instructionfollowing, alshikh2023selfinstruct}, which hinders their further applications. 

\begin{figure*}[th]
\centering 
\includegraphics[width=1.0\textwidth]{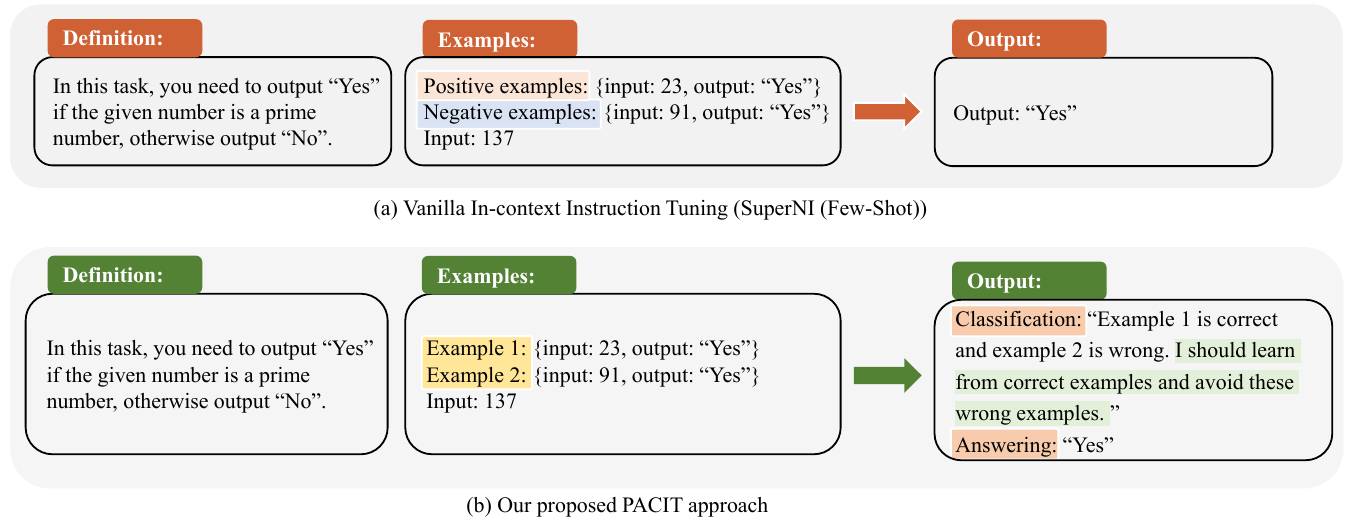}
\caption{The overview of \mnamec. \mname consists of two stages: Classification and Answering. (1) \textbf{Classification:} Judge the correctness of each provided example based on the task description and then take the self-reminder action (i.e., \textit{I should learn from correct examples and avoid wrong examples.}). (2) \textbf{Answering:} Respond to the main task instruction conditioned on the classification results. Two stages are executed sequentially within a single data sample.}
\label{fig:overview}
\end{figure*}

In this work, we introduce \mname, a simple and novel in-context instruction tuning approach (see Figure~\ref{fig:overview}) inspired by the pedagogical concept of \emph{desirable difficulty} \cite{wiki_desirable_difficulty,marshMemoryEducationalSettings2013a}. During finetuning with \mname method, the model first accomplishes a quiz about the judgment of correctness of the provided examples based on the task description, then responds to the task instance input. 
By transforming the provided example into a related quiz of the simple classification task, we encourage the model to be actively involved in recalling correlated information and grasping the distinction between positive and negative examples, going beyond surface-level information. In contrast to simply reading the examples, this approach enhances the model's comprehension of the task information, thereby improving its ability to follow instructions. 

Extensive experiments prove the effectiveness of \mname, outperforming ICIT baseline up to 9.16 and 3.14 average ROUGE-L \cite{lin-2004-rouge} on in-domain and out-of-domain datasets of SuperNI \citep{wang2022supernaturalinstructions}, respectively. The \mname still consistently surpasses traditional methods when the positive and negative examples are synthesized with self-instruct \citep{wang2023selfinstruct} by ChatGPT \cite{chatgpt}. Therefore, in cases that the human-crafted positive and negative examples are not available, the \mname has the potential to be a better instruction tuning strategy even for a large-scale instruction dataset.
Our contributions are summarized as follows:
\begin{itemize}
    \item We propose \mname, a simple yet effective in-context instruction tuning method that achieves better instruction following ability by better grasping the differences between positive and negative examples. 
    \item Extensive experiments demonstrate the superior performance of \mname over competitive baselines consistently across in-domain and out-domain datasets.
    \item The \mname also achieves better performance than vanilla instruction tuning when the examples are all synthesized with the self-instruct method.\footnote{The code is available at \url{https://github.com/XueTianci/PACIT}.}
\end{itemize}

\section{Related Work}
\subsection{Instruction Tuning}

Instruction tuning \cite{ouyang2022training} finetunes the pretrained language models with supervised instruction data to enhance the instruction following ability and enable the zero-shot generalization to unseen tasks \citep{chung2022scaling,wei2022finetuned,ouyang2022training,sanh2022multitask,alpaca}. The instruction tuning is an essential training stage for most large language models \cite{ouyang2022training,alpaca}. It commonly uses the next token prediction as the training objective. 

The key to instruction tuning is the quality and diversity of the instruction data \cite{zhou2023lima}. 
The instruction data used by InstructGPT \cite{ouyang2022training} is created with human experts. It can also be created with LLMs like ChatGPT \cite{chatgpt} with self-instruct \citep{wang2023selfinstruct} method. The self-instruct method synthesizes instruction data by prompting the LLM with few-shot examples and guidelines to use instructional signals from the model itself for data augmentation. The evol-instruct \citep{xu2023wizardlm} method further improves self-instruct to create more diverse instruction data with varying levels of complexities. The humpback \cite{liHumpBackSelfAlignmentInstruction2023} proposes to iterativly optimize the model and generate high-quality instruction data without the reliance on strong proprietary LLMs, similar to the back-translation practice in machine translation. 
Super natural instructions (SuperNI; \citealt{wang2022supernaturalinstructions}) is a benchmark that covers 76 distinct task types of 1616 diverse NLP tasks, including but not limited to classification, extraction, infilling, sequence tagging, text rewriting, and text composition. Each task in the SuperNI benchmark contains the task definition, task instances and example instances. Both task instance and example instance contain the input-output pairs for the task. The example instances have additional tags (i.e., positive or negative) based on the example and the task description. 

In-context instruction tuning \citep{ouyang2022training,wang2022supernaturalinstructions,li2023mimicit} finetune the LLMs with supervised instruction data as well as task-specific examples. The few-shot examples used in InstructGPT are all human-crafted positive examples.  \citet{wang2022supernaturalinstructions} further incorporates specified positive and negative crafted examples into the in-context instruction tuning. \citet{li2023mimicit} explore the in-context instruction tuning in the multimodal domain. Different from previous works that simply have the model passively read the examples, we explore to encourage the model to actively learn about the examples via verification the correctness of examples.

\subsection{In-Context Learning}

In-context learning (ICL; \citealt{liu-etal-2022-makes,rubin2022learning,min-etal-2022-metaicl}) is a prompt-based method that encourages the language models to learn from the few-shot examples presented in the model input. Researchers explore different approaches to improve the performance of ICL. 
\citet{min-etal-2022-metaicl} and \citet{chen2022metalearning} introduce meta-learning to better adapt the language models to ICL. \citet{zhao2021calibrate} estimates models' bias towards each answer and then develop contextual calibration to adjust the model's output probabilities. 
SG-ICL~\cite{kim2022selfgenerated} proposes to generate demonstration examples for in-context learning from the language model itself instead of humans. Active Prompting \citep{diao2023active} selects the most uncertain questions as demonstration examples to further improve the performance. \citet{min2022rethinking} finds that replacing gold labels with random labels only marginally hurts performance, which indicates models learn from the example format rather than input-label pairs. \citet{yoo2022groundtruth} revisit previous findings of \citet{min2022rethinking} and introduce novel metrics to prove that the input-label correspondence plays a more significant role in contextual demonstration than previously considered. However, most of these methods focus on the inference stage and explicitly show the correctness of the demonstration examples. Our work focuses on the instruction tuning stage.

\section{Method} \label{sec:method}

In this work, we focus on the in-context instruction tuning \cite{wang2022supernaturalinstructions} where both positive and negative examples are provided as the case in the SuperNI dataset (see Figure~\ref{fig:overview}). 
The model is trained to generate a response that is similar to the positive examples while avoiding the mistakes in the negative ones. 
Conventional works merely present these examples and their tags in the prompt following the practice of in-context learning. We propose \mname for better in-context instruction tuning by unlocking the power of provided examples. The \mname is motivated by the pedagogical psychological concept of \emph{desirable difficulty} \cite{marshMemoryEducationalSettings2013a,wiki_desirable_difficulty},
which improves the long-term performance of students by a learning task that requires a considerable but desirable amount of effort.

As an example of desirable difficulty, quizzing oneself with flashcards brings better learning outcomes than just reading the materials, as the quizzes require students to consistently recall associated information and encourage them to learn the material more concretely and actively. 
Simply reading the materials results in lower engagement and less attention from students. The key information and connected knowledge of the materials may be overlooked. In contrast, students think, analyze and try to apply their existing knowledge when they tackle a problem by hand. Active involvement in learning enhances their understanding of the knowledge, leading to better learning outcomes.
 
Following the insight of \emph{desirable difficulty}, the \mname proposes a supplementary quiz with the examples and asks the model to first accomplish the quiz before the task mentioned in the instruction. As shown in Figure~\ref{fig:overview}, the model is required to first classify the examples presented in the prompt into two types, positive or negative, according to the task description. The negative example indicates the unsatisfied output for the given input for this task, which should be avoided. After that, the model generates the response to the instruction based on the classification result of the provided examples. In this way, the model actively learns about the examples by accomplishing the related quiz, which further facilitates the understanding and grasp of the given task.

Consistent with SuperNI, each task has a task description $S_T$, a training dataset $\mathcal{D}=\{(X, Y)\}$, and an example pool consisting of positive and negative examples. For each input-output instance pair $(X, Y)$ in $\mathcal{D}$, we randomly select $k$ examples from the example pool and determine the order of positive and negative examples randomly.
Both the input and output of examples are presented in the prompt ($S^{in}_e = \{X_e, Y_e\}$), while the corresponding label $L_e$ (i.e., positive or negative) is set as the answer to the supplementary quiz and is part of the model output (see the example in Figure~\ref{fig:overview}). 
The ground-truth label of each example is replaced with the ordinal number and concealed in the input. In this way, the supplementary quiz is designed without human effort. 
Each data sample in \mname has two stages, i.e. \textbf{Classification} and \textbf{Answering}. 

\paragraph{Classification} The model is expected to judge the correctness of each provided example based on the task description during the classification stage. The ground-truth classification result $J_e$ is created from a template shown in Figure~\ref{fig:overview} and the example tag $L_e$. After giving the answer to the quiz, the model continues to generate the corresponding action to be taken $A_e$ (e.g., ``I should learn from correct examples and avoid mistakes in the wrong examples.''). The action serves as a self-reminder to encourage the model to take the corresponding action for better performance. 
During the first classification stage, the model is optimized with the next token prediction training objective. The ground-truth for action $A_e$ are human-crafted without tuning and kept the same for all samples. All tokens in the classification result and action are counted for the loss calculation. Formally, the loss of the classification stage can be represented as:
\begin{equation} \label{eq:stage1}
\mathbf{\mathcal{L}_{c}} = -\sum_{(X,Y) \in \mathcal{D}} \log P(J_e, A_e | S_T, S^{in}_e, X;\theta).
\end{equation}

\paragraph{Answering} 
Based on the result of the supplementary quiz $J_e$ and the corresponding action $A_e$, the model is elicited to output the answer $Y$ for instance input $X$ in the task. The answering stage is also trained with the language modeling objective. The corresponding training loss is calculated as
\begin{equation} \label{eq:stage2}
\hspace{-2mm}
\mathbf{\mathcal{L}_{a}}=\!-\!\sum_{(X,Y)\in \mathcal{D}}\!\log P(Y | S_T, S^{in}_e, X, J_e, A_e; \theta). 
\end{equation}

The overall training loss of \mname is the weighted sum of these two losses $\mathbf{\mathcal{L}} =\mathbf{\mathcal{L}_{c}} + \lambda \mathbf{\mathcal{L}_{a}}$, where $\lambda$ is a hyper-parameter to balance the two losses. During inference, the model generates the answer in the main task after completion of the auxiliary classification task.

\section{Experiments}
\subsection{Experiment Setting} \label{sec:setup}

\paragraph{Dataset}
We conduct experiments on the SuperNI-V2 dataset \citep{wang2022supernaturalinstructions}, an open-source dataset comprising over 800+ English tasks with diverse task types. Each task in the dataset includes four components: task definition, positive examples, negative examples and explanations. To ensure consistency, we utilize the same dataset split as SuperNI: the training set consisting of 756 diverse tasks and a hold-out test set containing 119 unseen out-domain tasks for evaluation purposes. Additionally, we construct a held-in test set that mirrors the training set's tasks but with different task instances to prevent any data leakage.
As the performance saturates when the number of instances per task increases \citep{wang2022supernaturalinstructions}, we randomly sample 60 instances for each task in the training set. For the test set, we randomly sample 100 instances for each task of the held-out test set and 15 instances for each task of the held-in test set, ensuring a comparable total number of instances for both datasets.

The statistics of our training, held-in and held-out datasets are presented in Table~\ref{tab:data_stat}.

\paragraph{Construction of Dataset.} To perform in-context instruction tuning, we construct the training dataset with data samples of the format \emph{task definition$+$positive/negative examples$+$task instance}. For each data sample, examples are added incrementally until the maximum input length is reached. Specifically, given a task instance, we first include the \emph{instance} and its corresponding \emph{task definition} to form a data sample. Subsequently, we randomly select a positive example and a negative example for the task and gradually add them to the data sample. To prevent the model from simply memorizing the corresponding tags, the order of the examples is shuffled. If adding an example exceeds the maximum input length limit, the addition process is stopped. This process results in four distinct types of data samples: (1) \textbf{Without examples}: training samples without any examples.
(2) \textbf{Only positive example}: training samples with only one positive example.
(3) \textbf{Only negative example}: training samples with only one negative example.
(4) \textbf{Mixing examples}: training samples with both positive and negative examples.
The proportions of these four types within our training data are 2.9\%, 6.3\%, 0.5\% and 90.2\%, respectively.
The few-shot inference dataset is constructed similarly, while the zero-shot inference dataset consists of data samples with the format \emph{task definition$+$task instance}.

\begin{table}[t]
\renewcommand{\arraystretch}{1.0}
\resizebox{\columnwidth}{!}{
\begin{tabular}{lccc}
\toprule
Statistics        & Train Set & Held-In & Held-Out \\ \midrule
Number of tasks             & 756               & 756     & 119      \\ 
\# of total instances   & 45360             & 11340    & 11900    \\ 
Avg. \# of Ex.    & 1.83              & 1.79    & 1.75     \\ \bottomrule
\end{tabular}}
\captionof{table}{Statistics of our training, held-in, and held-out datasets. `Avg. \# of Ex.' denotes the average number of examples per task. }
\label{tab:data_stat}
\end{table}

 \begin{table*}[th]
\renewcommand{\arraystretch}{1.0}
\resizebox{\linewidth}{!}
{
\begin{tabular}{cccccccc}
    \toprule
    \multirow{2}{*}{Model}   & \multirow{2}{*}{\begin{tabular}[c]{@{}c@{}}Testing Setting →\\ Training Setting ↓\end{tabular}} & \multicolumn{3}{c}{Held-Out}       & \multicolumn{3}{c}{Held-In}           \\ 
    \cmidrule(r){3-5}   \cmidrule(r){6-8}  &      & Zero-Shot       & Few-Shot        & Avg ROUGE-L       & Zero-Shot       & Few-Shot        & Avg ROUGE-L       \\ \midrule
    \multirow{3}{*}{\texttt{T5-770M}}   & SuperNI (Zero-Shot)     & \textbf{38.02}          & 40.59          & 39.30            & \textbf{46.22}           & 42.59          & 44.40          \\
                                        & SuperNI (Few-Shot)      & 33.30            & 45.08           & 39.19           & 43.59          & 52.96          & 48.27          \\
                                        & \mname                 & 33.59 & \textbf{46.66} & \textbf{40.13} & 44.67 & \textbf{53.31} & \textbf{48.99} \\ \midrule
    \multirow{3}{*}{\texttt{T5-3B}}    & SuperNI (Zero-Shot)      & 42.89           & 45.73          & 44.31          & \textbf{49.95}          & 47.59          & 48.77          \\
                                        & SuperNI (Few-Shot)      & 38.54          & 51.08          & 44.81          & 41.49           & 52.96          & 47.23          \\
                                        & \mname                 & \textbf{43.09}  & \textbf{52.11} & \textbf{47.60} & 47.29  & \textbf{55.21} & \textbf{51.25} \\ \midrule
    \multirow{3}{*}{\texttt{LLaMA2-7B}}    & SuperNI (Zero-Shot)     & 44.81           & 49.35          & 47.08          & 49.36          & 48.85          & 49.10          \\
                                        & SuperNI(Few-Shot)      & 42.14          & 50.71          & 46.43          & 45.53           & 52.68          & 49.10          \\
                                        & \mname                 & \textbf{45.62}  & \textbf{53.53} & \textbf{49.57} & \textbf{54.05}  & \textbf{62.47} & \textbf{58.26} \\ \bottomrule
    \end{tabular}
}
\captionof{table}{The comparison results of \mname and baselines under zero-shot and few-shot inference settings on hold-in and hold-out datasets. 
\textbf{Avg ROUGE-L:} we calculate the averaged ROUGE-L under zero-shot and few-shot inference settings. \textbf{Bold} denotes the best result.
}
\label{tab:main_results}
\end{table*}

\paragraph{Settings and Metrics}
Following \citet{kung2023models}, we utilize two variants of \texttt{T5-LM-Adapt} \cite{2020t5} as the backbones of \mname: \texttt{T5-Large-lm-adapt-770M} (T5-770M) and \texttt{T5-XL-lm-adapt-3B} (T5-3B). Additionally, to evaluate \mname with a stronger backbone, we conduct experiments using the \texttt{LLaMA-2-7B} (LLaMA2-7B) model. The $\lambda$ hyper-parameter is set as 1 when calculating the overall training loss.
During inference, we employ greedy decoding (i.e., set the temperature to 0) following \citet{wang2022supernaturalinstructions} to obtain the most confident predictions from the model outputs. 

Given the diversity of tasks and the open-ended generation nature of formulation, we adopt ROUGE-L metric \cite{lin-2004-rouge} for reporting aggregated performance results. The metric has been shown to correlate well with accuracy for classification tasks and human evaluation \citep{wang2022supernaturalinstructions}. Unless otherwise specified, we report results on the held-out dataset in the Ablation Study (Section \ref{sec:ablation_study}) and Analyses (Section \ref{sec:analyses}).

\paragraph{Training Details}
We use Adam optimizer with $\beta_1 = 0.9$, $\beta_2 = 0.999$ to finetune the models. The models are trained for five epochs and the last checkpoint is used for evaluation. The global batch size is 64. We use the linear learning rate scheduler. The learning rate for T5-based models is set to $2 \times 10^{-4}$ following \citet{kung2023models}, while the learning rate for LLaMA-2 is set to $2 \times 10^{-5}$ following \citet{alpaca,chen2023alpagasus}. We set the maximum input length as 1024 and the maximum output length as 128 for all models following \citet{wang2022supernaturalinstructions}.
All experiments are run on eight NVIDIA RTX-4090 GPUs using \texttt{Huggingface Transformers}\footnote{\url{https://github.com/huggingface/transformers}} toolkit.

\paragraph{Baselines}
We compare \mname with two baselines:
\begin{itemize}
    \item SuperNI (Zero-Shot): We formulate each data sample as \emph{task definition$+$main task instance} and train with conventionally instruction tuning method. No examples are used during training for this setup. 
    \item SuperNI (Few-Shot): We use the same training dataset as \mname, but train with conventionally in-context instruction tuning. In the subsequent text, we may use SuperNI to denote this method for simplicity.
\end{itemize}

\subsection{Main Results}  \label{sec:exp_results}

To assess the efficacy of \mnamec, we compare it with baselines as presented in Table~\ref{tab:main_results}. As can be seen, \mname consistently outperforms SuperNI (Zero-Shot) and SuperNI (Few-Shot) methods across the held-in and held-out datasets. Notably, the performance gap is more pronounced for larger models compared to smaller model. Specifically, when utilizing the T5-3B and LLaMa2-7B models, \mname exhibits substantial improvements over the SuperNI (Few-Shot) method, with average ROUGE-L score boosts of 2.79 and 3.14 on the held-out test set, and 4.02 and 9.16 on the held-in test set, respectively. Conversely, smaller T5-770M model demonstrates only marginal increases of 0.94 and 0.72 average ROUGE-L scores.
We hypothesize that larger models, which have stronger learning capabilities, can excavate more internal information in demonstration examples with our proposed \mname methods. 
Additionally, it is noteworthy that \mname exhibits greater improvements on the held-in datasets compared to the held-out datasets, indicating its ability to significantly benefit seen tasks.
In the zero-shot inference setting, SuperNI (Zero-Shot) method achieves good performance. However, its performance sharply declines in the few-shot setting. This discrepancy can be attributed to the importance of maintaining consistency between the training and inference settings.

To further showcase the effectiveness of PACIT, we also assess its performance in the MMLU benchmark \citep{hendrycks2020measuring}, which includes 57 subjects at varying difficulty levels with a multiple-choice format. The results are shown in table \ref{tab:MMLU_results}. We can observe that PACIT significantly enhances performance in both zero-shot and few-shot scenarios, boosting accuracy by 5.59\% and 4.46\%. In summary, \mname outperforms all baselines and achieves new state-of-the-art on ICIT.

\begin{table}[h]
\centering
\renewcommand{\arraystretch}{0.6}
\resizebox{0.9\columnwidth}{!}{
\begin{tabular}{lccc}
\toprule
Method         & Zero-Shot & Few-Shot & Avg. \\ \midrule
Base           &28.67\%	&45.30\%  &36.99\% \\
SuperNI         & 44.02\% & 46.76\% & 45.39\% \\ 
\mname         & \textbf{49.61\%}  & \textbf{51.22\%} & \textbf{50.42\%}          \\ 
$\Delta$ (\%)   & \textcolor{darkpastelgreen}{+5.59}         & \textcolor{darkpastelgreen}{+4.46}          & \textcolor{darkpastelgreen}{+5.03}          \\ 
\bottomrule
\end{tabular}}
\captionof{table}{The comparison results of Pacit and baselines in the MMLU benchmark. \textbf{Base:} The performance of the original LLaMA-2 model. For the few-shot setting, we use 5-shot as previous works \citep{hendrycks2020measuring,fu2023chain}.}
\label{tab:MMLU_results}
\end{table}

\begin{figure*}[t]
\begin{minipage}{0.49\textwidth}
     \centering
    \resizebox{\textwidth}{!}{
    \includegraphics{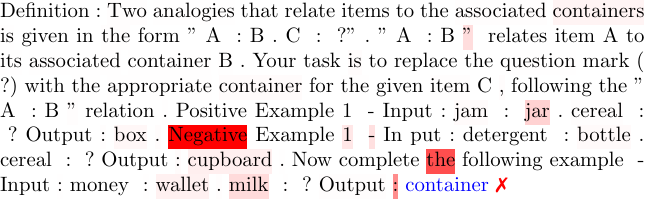}
    }
    \centerline{(a) SuperNI (Few-Shot)}
\end{minipage}
\hfill
\begin{minipage}{0.49\textwidth}
    \centering
    \resizebox{\textwidth}{!}{
    \includegraphics{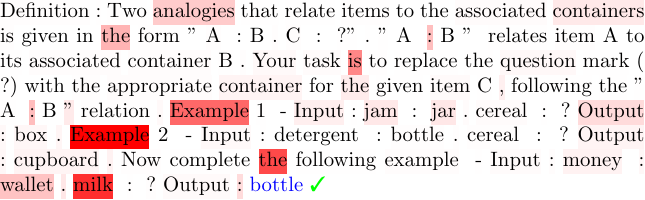}
    }
    \centerline{(b) \mname}
\end{minipage}
\caption{A concrete example of attention visualization for SuperNI (Few-Shot) and \mname methods.} 
\label{fig:vis}
\end{figure*}

\subsection{Ablation Study} \label{sec:ablation_study} 
We conduct an ablation study on the training method of \mnamec. Initially, we begin with \mnamec, which consists of two training stages: classification with action, and answering. Subsequently, we gradually remove the action after classification (setting (2)) and the whole classification stage to roll back to the vanilla SuperNI (Few-Shot) method (setting (3)). 
\begin{table}[h!]
\centering
\resizebox{0.9\columnwidth}{!}{
\begin{tabular}{llccc}
\toprule
ID & Method         & ZS & FS & Avg. \\ \midrule
(1) &\mname         & \textbf{43.09} & \textbf{52.11} & \textbf{47.60} \\ 
(2) & (1)$-$action & 41.48         & 51.29          & 46.38          \\ 
(3)&(2)$-$aux.   & 38.50         & 51.08          & 45.15          \\ 
(4)&(3)$+$separate aux. (w. action)  & 39.87         & 51.23          & 45.55 \\ 
\bottomrule
\end{tabular}}
\captionof{table}{The performance (ROUGE-L) of ablation study variants (ZS=zero-shot inference, FS=few-shot inference) on held-out set. Starting from \mname, we gradually remove the action (ID=2) and the auxiliary classification stage (aux., ID=3) in each data sample. }
\label{tab:ablation_result}
\end{table}
To further explore the necessity of integrating classification and answering within a single data sample, we separate a standard \mname training data sample into two sub-samples: a SuperNI (Few-Shot) data sample and a classification sample of provided examples (i.e., judge whether these examples satisfy the requirements of task definition and then generate action) (setting (4)). This setting corresponds to multi-task learning, where the model is jointly trained with data samples from different tasks. An illustrative classification sample is provided in Figure~\ref{app:template_separate} of the appendix. It's worth noting that we do not perform classification tasks for each example but combine multiple examples together for classification. This strategy mitigates performance variations that may arise from disparate training task proportions.

The results are shown in Table~\ref{tab:ablation_result}. Removing the action leads to a decrease of 1.22 average ROUGE-L score, and further removing the classification stage results in an additional decrease of 1.23 average ROUGE-L score. This observation confirms our insights regarding \emph{desirable difficulty}, as the inclusion of a supplementary quiz on the examples and an action to emphasize its importance guides the model to enhance its learning from the examples.
Furthermore, when comparing setting (1) and setting (4), we find that it is necessary to integrate classification and answering within a single data sample, as seperating them reduces performance by 2.05 average ROUGE-L score.

\section{Analyses} \label{sec:analyses}

\paragraph{The Visualization of Attention.}
To better understand how \mname works, we conduct a case study by visualizing the attention weights in T5-3B model. We visualize the averaged encoder-decoder attention weights of different heads in the last layer of T5-3B. Figure \ref{fig:vis} shows a concrete example of \mname v.s. SuperNI (Few-Shot). The color in each figure represents the relative attention weights. Actually, the relative attention weights is also based on the generated classification results and actions for PACIT. In order to show the comparison more clearly, we do not include them in the figure.
As can be seen, \mname allocates more attention to the task definition and examples' information compared with the SuperNI (Few-Shot) model. The attention weights from \mname exhibit a broader span across the prompt. This observation is expected as the classification task in \mname encourages the model to focus more on task definition and examples, otherwise it cannot classify examples correctly. We also manually check some other examples which present similar patterns. 

\begin{table}[h]
\centering
\renewcommand{\arraystretch}{1.0}
\resizebox{\columnwidth}{!}{
\begin{tabular}{lcccc}
    \toprule
    Benchmark & Method & Zero-Shot         & Few-Shot & Input/Output tokens\\ \midrule
    \multirow{3}{*}{SuperNI}    & SuperNI     & 42.14        & 50.71         & 542/\textbf{20}         \\ 
                                & SuperNI+SC     & 41.48        & 50.17      & 2750/95                 \\
                                & \mname               & \textbf{45.62}       & \textbf{53.53}     & \textbf{540}/48    \\ \midrule
    \multirow{3}{*}{MMLU}      & SuperNI  & 44.02\%  & 46.76\%   & N/1    \\ 
                                & SuperNI+SC     & 44.37\%   &47.00\%  & 5N/1       \\
                                & \mname  & \textbf{49.61} & \textbf{51.22}   & N/1    \\ 
    \bottomrule
    \end{tabular}}
\caption{The performance with additional computation tokens by Self-Consistency in zero-shot and few-shot inference settings.}
\label{tab:sc}
\end{table}

\paragraph{The Effectiveness of Additional Computation}
Considering that the effectiveness of \mname may come from trading off the extra token compute, we also compare \mname with the Self-Consistency\citep{wang2023selfconsistency}(SC for short) method in the inference stage based on LLaMA-2 model to maintain similar computational overhead. Specifically, we conducted 5 trials per problem for SC with temperature 0.7\citep{wang2023selfconsistency}. More sampling results at different temperatures can be found in the appendix \ref{app:temperatures}.

Table \ref{tab:sc} shows the results of additional computation tokens in MMLU and SuperNI benchmarks. In the SuperNI benchmark, we can observe that when allocating additional inference computation through the SC method leads to a decline in performance. This is because SuperNI tasks are more open-ended (such as QA and translation), which do not have fixed answers like arithmetic or logical reasoning. Even after sampling five times, there are five possible different answers. Therefore, combining these answers through SC does not yield performance improvements. On the contrary,  A mere increase of 28 tokens per question in the output by \mname results in a substantial performance improvement, yielding a 3.48 and 2.8 Rouge-L score improvement in zero-shot and few-shot settings, respectively. In the MMLU tasks, allocating additional inference computation through the SC method can improve the performance, gaining 0.35\% and 0.24\% in zero-shot and few-shot settings. However, we can also observe that PACIT still significantly outperforms Super+SC method with the same token cost as standard SuperNI and five times lower compared to Super+SC(improving the accuracy by 5.59\% and 4.46\% in MMLU in zero-shot and few-shot settings, respectively.). In summary, the performance gain achieved by PACIT is primarily attributable to quizzing rather than solely allocating more computation. It is also worth noting that PACIT has the same cost as the standard SuperNI. This is because MMLU doesn't contain the corresponding format as the SuperNI dataset (such as task definition), and PACIT just directly answers the choice for problems in a common way without classification, which suggests PACIT may not necessarily need to be consistent with the training format to bring improvement, showing generalization ability. This also further demonstrates that the effectiveness of PACIT comes from the training stage by additional auxiliary classification tasks rather than the inference stage.

\begin{figure}
\centering
\includegraphics[width=0.8\columnwidth]{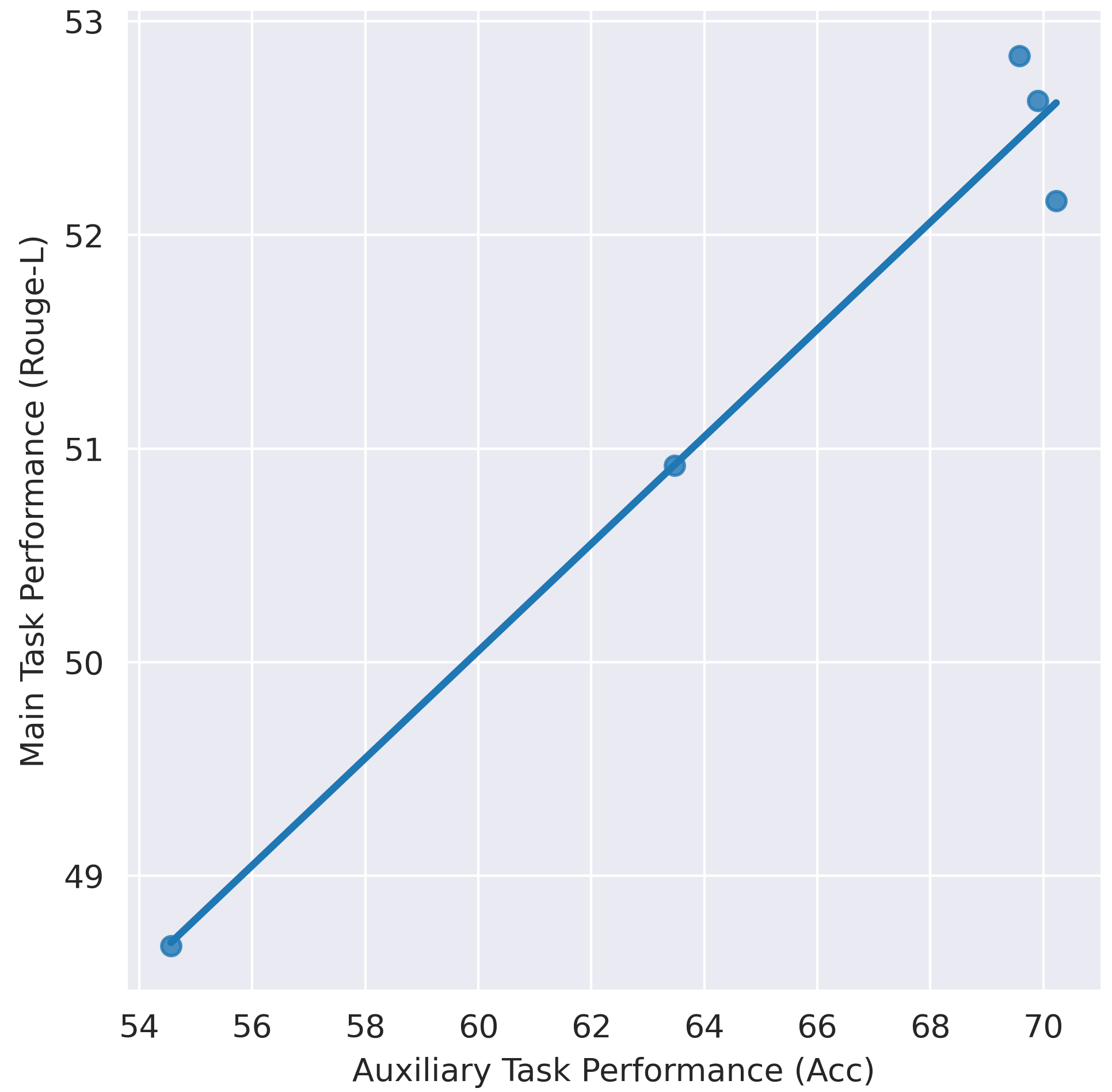}
\caption{The training dynamics of the main task (ROUGE-L) v.s. the auxiliary classification task (Acc). \textbf{Acc:} The accuracy of classification. \textbf{ROUGE-L:} The performance of main tasks. The five data points represent five checkpoints obtained after each epoch.}
\label{fig:scatter}
\end{figure}

\paragraph{The Relationship between Classification Accuracy and Model Performance.}
To gain insights into the correlation between the auxiliary task (i.e., classification) and main task, we analyze the training dynamics by plotting the main task's performance (ROUGE-L) against the auxiliary task's performance (Acc). The results are shown in Figure~\ref{fig:scatter}. 
The classification accuracy demonstrates a strong correlation with the main task's ROUGE-L score, as evidenced by the slope. Furthermore, we calculate the Pearson correlation coefficient between these two metrics, resulting in a high value of 0.98.
While correlation does not establish causation, it does provide valuable insights into the interpretability of \mname.

\begin{table*}[th]
\renewcommand{\arraystretch}{1.0}
\centering
\resizebox{0.85\linewidth}{!}{
\begin{tabular}{lccccc}
    \toprule
    \multirow{2}{*}{Model}    & Testing Setting $\rightarrow$   & \multirow{2}{*}{Zero-Shot}  &  \multicolumn{3}{c}{Few-Shot}  \\
              \cmidrule(r){4-6}                         & Training Setting $\downarrow$  &      & Generated    & Ground-Truth     &  Random                        \\ \midrule
    \multirow{4}{*}{\texttt{T5-770M}}  & SuperNI (Ground-Truth)  &33.30 &- &45.08 &45.26  \\
                    & SuperNI (Random)  &30.66 &- &43.54 &43.48  \\
                    & \mname(Ground-Truth)      & 33.58      & \textbf{46.66}     & \textbf{46.67}         & \textbf{46.72}        \\
                    & \mname(Random)            & \textbf{34.23}      & 46.17    & 46.10          & 46.11       \\ \midrule
    \multirow{4}{*}{\texttt{T5-3B}}    & SuperNI (Ground-Truth)  &38.54 &- &51.08 & 51.25\\
                & SuperNI (Random)  &36.71 &- &49.12 &48.92   \\ 
                & \mname(Ground-Truth)     & \textbf{43.09}      & \textbf{52.11}     & \textbf{52.17}   & \textbf{52.07}  \\
                & \mname(Random)             & 33.52     & 45.76     & 46.14      & 46.11  \\
    \bottomrule
    \end{tabular}
}
\captionof{table}{The Performance (ROUGE-L) on held-out set with different classification labels in the training and inference time. 
We compare two training settings and three inference settings for the labels of few-shot examples in each data sample. \textbf{Generated}: classification labels generated from the model; \textbf{Ground-Truth}: true classification labels; \textbf{Random}: randomly sampled classification labels. 
}
\label{tab:influence_of_label}
\end{table*}

\paragraph{The Effect of Classification Labels in Training and Inference Phase.}
Inspired by previous work on in-context learning \citep{min2022rethinking, madaan-etal-2023-makes, wei2023larger}, we suspect \mname utilize examples either by (a) recognizing the task from examples and applying LLMs’ pre-trained priors (learning the format \citep{min2022rethinking}) and/or (b) learn the input–label mappings from the presented examples (learning the input-label mapping). When ground-truth labels are provided during in-context instruction tuning, these two factors operate simultaneously. To study which of these factors drives performance, we compare two training settings: 
\begin{itemize}[]
    \item \textbf{Ground-Truth:} The true classification labels are used, which is the standard setup of \mnamec.
    \item \textbf{Random:} The classification labels are uniformly sampled from the label space. In this setup, LLMs can only learn the format.
\end{itemize} 

\begin{table*}[h]
\renewcommand{\arraystretch}{1.0}
\centering
\resizebox{0.85\textwidth}{!}{
\begin{tabular}{lccccc}
    \toprule
    \multirow{2}{*}{Model} & \multirow{2}{*}{\begin{tabular}[c]{@{}c@{}}Testing Setting →\\ Training Setting ↓\end{tabular}} & \multirow{2}{*}{Zero-Shot} & \multirow{2}{*}{Few-Shot} & \multirow{2}{*}{Avg. ROUGE-L} \\ 
    \multicolumn{1}{c}{}                       & \multicolumn{1}{c}{}      &   & \multicolumn{1}{l}{}      &                           &                             \\ \midrule 
    \multirow{4}{*}{T5-770M}  & SuperNI (1 pos and 1 neg)      & 33.30                      & 45.08                     & 39.19                       \\
                              & SuperNI (2 pos and 2 neg)      & 30.75                    & 45.82                    & 38.28                      \\
                              & \mname (1 pos and 1 neg)       & \textbf{33.59}           & \textbf{46.66}           & \textbf{40.13}             \\
                              & \mname (2 pos and 2 neg)       & 28.66                    & 45.85                    & 37.26                      \\ \midrule 
    \multirow{4}{*}{T5-3B}    & SuperNI (1 pos and 1 neg)      & 38.54                    & 51.08                    & 44.81                      \\
                              & SuperNI (2 pos and 2 neg)      & 35.72                    & 49.64                    & 42.68                       \\
                              & \mname (1 pos and 1 neg)       & \textbf{43.09}            & \textbf{52.11}           & \textbf{47.60}             \\
                              & \mname (2 pos and 2 neg)       & 38.92                    & 51.41                    & 45.17                      \\ \bottomrule
    \end{tabular}
}
\caption{The performance (ROUGE-L) on held-out set with different numbers of demonstration examples in zero-shot and few-shot inference settings. \textbf{N pos and M neg:} There are N positive examples and M negative examples in each training sample at most.}
\label{tab:num_examples}
\end{table*}

Table~\ref{tab:influence_of_label} shows the results. At the inference stage, in addition to the standard inference setup of \mname that generates classification labels from the model (\textbf{Generated}), we also explore Ground-Truth and Random variants. 
As can be seen, \mname with Ground-Truth training setting exhibits a significantly greater improvement over Random training setting on large model (T5-3B) compared to small model (T5-770M). This observation reminds us  previous research on in-context learning, which suggests that \textbf{learning the format is a broader capability across
scales, while learning the input-label mapping is enabled with scale \cite{wei2023larger,pan-etal-2023-context,kossen2023incontext}}. We speculate that large model is better at learning input-output mapping than small model for \mnamec. 
When comparing different inference setups, we find that the model tuned by \mname is insensitive to labels at the inference stage for both small and large models. This aligns with previous work's \citep{wei2023larger} observation that instruction-tuned models rely more on their own semantic priors so that they are less influenced by the labels presented in examples when conducting few-shot inference. For SuperNI, we find that random labels at training time influence small and big models similarly. We leave more in-depth studies as future work. 

\paragraph{The Influence of Number of Demonstration Examples.}

Humans can improve their ability to complete downstream tasks by learning from more demonstration examples. Therefore, we construct experiments to explore whether more examples in each data sample lead to better performance. Since the average number of positive examples and negative examples of the SuperNI dataset are 2.8 and 2.4, we cannot conduct experiments with a maximum number of examples greater than 3. The results are shown in Table \ref{tab:num_examples}. We use the same number of demonstration examples in both training and few-shot inference time. Overall, more examples consistently lead to performance degradation for both SuperNI and \mname in zero-shot and few-shot settings. For example, the performance of \mname on T5-770M and T5-3B drops by 2.86 and 2.43 average ROUGE-L when switching from a pair of positive and negative examples to two pairs, respectively. We suspect with more demonstration examples, \mname as well as SuperNI could be misguided by interference among examples and their spurious correlations. A similar phenomenon has been observed in in-context learning. We refer the readers to \citet{chen2023demonstrations} for more detailed discussions.

\paragraph{The Performance of \mname with Generated Examples.}
\begin{table*}[h]
\renewcommand{\arraystretch}{1.0}
\centering
\resizebox{0.75\textwidth}{!}{
\begin{tabular}{lcccc}
    \toprule
    \multirow{2}{*}{Model} & \multirow{2}{*}{\begin{tabular}[c]{@{}c@{}}Testing Setting →\\ Training Setting ↓\end{tabular}} & \multirow{2}{*}{Zero-Shot}         & \multirow{2}{*}{Few-Shot} & \multirow{2}{*}{Avg ROUGE-L} \\
                                &                                   &                                    &                           &                             \\ \midrule
    \multirow{3}{*}{T5-770M}    & SuperNI (Zero-Shot)     & \textbf{32.66}        & 37.50         & 35.08         \\ 
                                & SuperNI (Few-Shot)     & 23.08        & 40.54      & 31.81                 \\
                                & \mname               & 32.62       & \textbf{41.16}     & \textbf{36.89}      \\ \midrule
    \multirow{3}{*}{T5-3B}      & SuperNI (Zero-Shot)  & 37.63  & 41.53   & 39.58       \\ 
                                & SuperNI (Few-Shot)     & 36.38     & 43.09      & 39.73       \\
                                & \mname  & \textbf{37.95} & \textbf{44.23}   & \textbf{41.09}    \\ 
    \bottomrule
    \end{tabular}}
\caption{The Performance (ROUGE-L) with generated examples (by Self-Instruct) in zero-shot and few-shot inference settings.}
\label{tab:self_instruct_result}
\end{table*}
A limitation of \mname is its reliance on positive and negative examples during training. However, the positive and negative examples are not readily available for many instruction datasets. As human annotation is expensive and time-consuming, we tackle the problem by leveraging automatically generated examples from LLM. Specifically, we generate examples with the self-instruct \citep{wang2023selfinstruct} method, which is a framework for improving the instruction-following capabilities of LLMs by bootstrapping off their own generations. We choose the ChatGPT (\texttt{gpt-3.5-turbo-0613}) as the backbone LLM and set the temperature to 0.7 to improve the diversity of generated data. To create our example seed pool, we randomly select eight pairs of positive and negative examples in total from all examples of different tasks.
For each generation, we construct the prompt with task definition and few-shot demonstrations to generate new pairs of positive and negative examples. The few-shot demonstrations consist of four pairs of positive and negative examples and their corresponding task definitions randomly sampled from the seed pool. In this way, we reduce the number of annotated training examples from 1384 to 8.
Due to the API expense of the proprietary LLM, we only construct 5040 training samples (84 different tasks with 60 training samples each). The entire data template for generating new positive and negative examples is shown in the appendix~\ref{app:template} (see Figure~\ref{fig:app_template_self_instruct}). 

The performance with generated examples is shown in Table~\ref{tab:self_instruct_result}. As can be seen, with generated examples, \mname improves over baseline without any examples (SuperNI (Zero-Shot)) by 1.81 Avg ROUGE-L on T5-770M and 1.51 Avg ROUGE-L on T5-3B, and vanilla in-context instruction tuning baseline (SuperNI (Few-Shot)) by 5.08 Avg ROUGE-L on T5-770M and 1.63 Avg ROUGE-L on T5-3B. These results are particularly impressive considering that the quantity of our samples accounts for only 11\% of the samples used in the main experiment and the generated examples from self-instruct are noisy \citep{wang2023selfinstruct}. Furthermore, we find that the improvement brought by \mname over SuperNI (Zero-Shot) is larger for T5-770B compared with T5-3B. This finding contrasts with the main experiments, where T5-3B exhibits an additional 2.46 average ROUGE-L improvement over T5-770M. This disparity can be attributed to small model's limited ability to learn from the input-label mapping, as its performance is less affected by noisy labels generated by self-instruct. 

\section{Conclusions}
In this paper, we introduce \mnamec, an effective in-context instruction tuning approach that unlocks the power of examples to enhance the instruction following ability of LLMs. Inspired by the pedagogical observations, \mname proposes to encourage the model to actively learn and comprehend the differences between the provided positive and negative examples rather than passively reading them. The model completes a quiz to assess the correctness of examples first and subsequently responds to the main task instruction based on the grasp of the examples. Experiments on SuperNI dataset demonstrate the superior performance of \mname over competitive baselines.  
In our preliminary experiment, \mname is observed to improve the performance of instruction tuning with positive and negative examples created with the self-instruct method, which shows a promising approach for better instruction tuning with large-scale instruction data. However, the generated examples with self-instruct method need further filtering to enhance the performance of \mname as the noisy examples may have negative impact on the performance. We leave the exploration of filtering the augmented data as well as scaling \mname to larger models like LLaMA-2-13B, LLaMA-2-70B and larger datasets as future work. 

\section{Limitations}

The proposed \mname method requires both positive and negative examples which are not readily available for many instruction datasets. These examples can be created with human efforts, resulting in additional expenses. They can also be synthesized with self-instruct method or other LLM-based data augmentation methods. In this case, the generated data samples need to undergo additional filtering following the common practice of data augmentation.

\subsection*{Acknowledgments}

This project was supported by National Natural Science Foundation of China (No. 62306132, No. 62106138). We thank the anonymous reviewers for their insightful feedbacks on this work.

\bibliography{conference}

\appendix
\section{Data Templates}
\label{app:template}

\paragraph{1. Data Template for \mnamec.} Our proposed \mname method takes the task definition, examples and instance input as the prompt. The model first generates the response to the auxiliary classification task and corresponding action of the provided examples. Based on the quiz result and action to be taken, the model then produces the outputs for the instance input for the given task. 

\begin{tcolorbox}[fonttitle=\bfseries]
\small{
\textbf{Task Definition: \textcolor{blue}{\{\{definition\}\}}}\\
\textbf{Example 1 }\\-
\hspace{0.2cm}\textbf{Input: \textcolor{blue}{\{\{exp.input\}\}}}\\-
\hspace{0.2cm}\textbf{Output: \textcolor{blue}{\{\{exp.output\}\}}}\\
\textbf{Example 2 }\\-
\hspace{0.2cm}\textbf{Input: \textcolor{blue}{\{\{exp.input\}\}}}\\-
\hspace{0.2cm}\textbf{Output: \textcolor{blue}{\{\{exp.output\}\}}} \\
\textbf{Evaluation Instance }\\-
\hspace{0.2cm}\textbf{Input: \textcolor{blue}{\{\{exp.input\}\}}}}
\tcblower
\small{
\textbf{Classification }\\-
\hspace{0.2cm}\textbf{Classification result: \textcolor{blue}{\{\{Example 1 is correct/wrong and example 2 is correct/wrong.\}\}}}\\-
\hspace{0.2cm}\textbf{Generated action: \textcolor{blue}{\{\{I should learn from correct examples and avoid the mistakes in these wrong examples.\}\}}}}
\tcbline
\small{
\textbf{Answering }\\-
\hspace{0.2cm}\textbf{Output: \textcolor{blue}{\{\{exp.output\}\}}}
}
\end{tcolorbox}
\captionof{figure}{The data template used for \mname method.}
\label{template_TADIS}

\paragraph{2. Data Template Used when Separating Two Stages of \mnamec.} The data template shown in Figure~\ref{app:template_separate} is used when the model is trained with separated classification and few-shot answering in Section~\ref{sec:ablation_study}. In this case, the model will only verify the correctness of provided examples in the classification sub-task instead of one stage of \mnamec. 

\begin{tcolorbox}[fonttitle=\bfseries]
\small{
\textbf{Task Definition: \textcolor{blue}{\{\{definition\}\}}}\\
\textbf{Example 1 }\\-
\hspace{0.2cm}\textbf{Input: \textcolor{blue}{\{\{exp.input\}\}}}\\-
\hspace{0.2cm}\textbf{Output: \textcolor{blue}{\{\{exp.output\}\}}}\\
\textbf{Example 2 }\\-
\hspace{0.2cm}\textbf{Input: \textcolor{blue}{\{\{exp.input\}\}}}\\-
\hspace{0.2cm}\textbf{Output: \textcolor{blue}{\{\{exp.output\}\}}}\\
\textbf{Judge whether each example conforms to the task definition.}
\tcbline
-\hspace{0.2cm}\textbf{Prediction: \textcolor{blue}{\{\{Example 1 is correct/wrong and example 2 is correct/wrong. I should learn from correct examples and avoid mistakes in the wrong examples.\}\}}}}
\end{tcolorbox}
\label{app:template_separate}
\captionof{figure}{The data template used for the classification task when training with separated two stages.}

\paragraph{2. Data Template for Generating Examples with Self-Instruct.} When generating positive and negative examples with the Self-instruct method, we randomly select four pairs of positive and negative examples in total from all examples of different tasks in the SuperNI dataset as in-context learning examples. We use ChatGPT (gpt-3.5-0613) to generate a positive and negative example pair based on the prompt shown in Figure~\ref{fig:app_template_self_instruct}.

\begin{tcolorbox}[fonttitle=\bfseries]
\small{
\textbf{Few-Shot Demonstrations:}\\
\textbf{Demonstrated Task Definition: \textcolor{blue}{\{\{definition\}\}}}\\
\textbf{Positive Example}\\-
\hspace{0.2cm} \textbf{Input: \textcolor{blue}{\{\{exp.input\}\}}}\\-
\hspace{0.2cm} \textbf{Output: \textcolor{blue}{\{\{exp.output\}\}}}\\
\textbf{Negative Example}\\-
\hspace{0.2cm} \textbf{Input: \textcolor{blue}{\{\{exp.input\}\}}}\\-
\hspace{0.2cm} \textbf{Output: \textcolor{blue}{\{\{exp.output\}\}}}\\
\textbf{......} \\
\textbf{Generated Examples:}\\
\textbf{Task Definition: \textcolor{blue}{\{\{definition\}\}}}
\tcbline
\textbf{Positive Example}\\-
\hspace{0.2cm} \textbf{Input: \textcolor{blue}{\{\{gen.input\}\}}}\\-
\hspace{0.2cm} \textbf{Output: \textcolor{blue}{\{\{gen.output\}\}}}\\
\textbf{Negative Example}\\-
\hspace{0.2cm} \textbf{Input: \textcolor{blue}{\{\{gen.input\}\}}}\\-
\hspace{0.2cm} \textbf{Output: \textcolor{blue}{\{\{gen.output\}\}}}}
\end{tcolorbox}
\captionof{figure}{The data template for generating positive and negative examples with the Self-instruct method.}
\label{fig:app_template_self_instruct}

\section{Different sampling temperatures}\label{app:temperatures}
Table \ref{tab:sc_temperature} shows the influence of different sampling temperatures on the SC results. Due to the open-ended format of the SuperNI task, sampling multiple times and selecting the most consistent answers will not improve performance and even lead to a slight decrease. Additionally, as the sampling temperature increases, performance decreases further. This may be due to increased diversity leading to uncertainty in the answers.

\begin{table}[h]
\centering
\resizebox{\columnwidth}{!}{
\begin{tabular}{lccc}
    \toprule
    Temperature & Zero-Shot        & Few-Shot & Avg \\ \midrule
    0.0    & 42.14\%        & 50.71\%         & 46.43\%          \\ 
    0.3    & 41.92\%        & 50.70\%      & 46.31\%                  \\
    0.5   & 41.75\%       & 50.49\%     & 46.12\%      \\
    0.7   & 41.48\%  & 50.17\%   & 45.83\%      \\ 
    PACIT & \textbf{45.62\%}   &\textbf{53.53\%}  & \textbf{49.58\%}       \\
    \bottomrule
    \end{tabular}}
\caption{The performance with Self-Consistency at different temperatures in zero-shot and few-shot inference settings.}
\label{tab:sc_temperature}
\end{table}

\end{document}